\documentclass[11pt]{article}

\usepackage[preprint]{acl}

\usepackage{times}
\usepackage{latexsym}

\usepackage[T1]{fontenc}

\usepackage[utf8]{inputenc}

\usepackage{microtype}

\usepackage{inconsolata}

\usepackage{graphicx}

\usepackage{amsmath}
\usepackage{amssymb}
\usepackage{nameref}
\usepackage{graphicx} 
\usepackage{booktabs}
\usepackage{multirow}
\usepackage{array}
\usepackage{makecell}
\usepackage[table,dvipsnames]{xcolor}
\definecolor{babyblue}{HTML}{DAEFF9}
\usepackage{colortbl}
\newcolumntype{C}[1]{>{\centering\arraybackslash}m{#1}}
\usepackage{xcolor}
\usepackage[most]{tcolorbox}

\newtcbtheorem[
  auto counter,
  number within=section,
  list inside=prompt
]{prompt}{Prompt}{
  enhanced,
  breakable,
  colbacktitle=black!60,
  coltitle=white,
  fontupper=\footnotesize,
  boxsep=5pt,
  left=0pt,right=0pt,top=0pt,bottom=0pt,
  boxrule=1pt
}{pr}

\newcommand{\methodname}{ToolScope}

\title{ToolScope: An Agentic Framework for Vision-Guided and Long-Horizon Tool Use}

\author{
  Mengjie Deng \quad Guanting Dong \quad Zhicheng Dou$*$ \\
  Gaoling School of Artificial Intelligence, Renmin University of China \\
  \texttt{\{mengjiedeng, dou\}@ruc.edu.cn}
}

\begin{document}
\maketitle
\begin{abstract}
Recently, large language models (LLMs) have demonstrated remarkable problem-solving capabilities by autonomously integrating with external tools for collaborative reasoning. However, due to the inherently complex and diverse nature of multimodal information, enabling multimodal large language models (MLLMs) to flexibly and efficiently utilize external tools during reasoning remains an underexplored challenge.
In this work, we introduce \textbf{\methodname}, an agentic framework designed to unify global planning with local multimodal perception, adopting a specialized Perceive tool to mitigates visual context degradation in long-horizon VQA task. \methodname\ comprises three primary components: the Global Navigator, the Agentic Executor, and the Response Synthesizer. The \textbf{Global Navigator} functions as a "telescope", offering high-level strategic guidance. The \textbf{Agentic Executor} operates iteratively to augment MLLM with local perception through the integration of external tools-Search, Code, and Perceive. Finally, the \textbf{Response Synthesizer} consolidates and organizes the reasoning process into a coherent, user-friendly output.
We evaluate \methodname\ on four VQA benchmarks across diverse domains, including VQA 2.0, ScienceQA, MAT-Search and MathVista. It demonstrates strong generalization capabilities, achieving an average performance improvement of up to +6.69\% across all datasets.
\end{abstract}

\section{Introduction}

Recent advances in Multimodal Large Language Models (MLLMs), such as GPT-4V~\citep{GPT-4V}, Gemini~\citep{Gemini}, and Claude~\citep{Claude3}, have led to impressive performance on a wide range of vision-language tasks. However, these models often struggle to adapt to more complex task that require dynamic reasoning, external knowledge access, or multi-step computation.
To address these limitations, multimodal agents~\citep{Magma, AnyMAL, HYDRA, Chain-of-Image} have emerged as a promising paradigm. By enabling MLLMs to interact with external tools, these agents extend the model's capabilities beyond the scope of end-to-end model inference. Hence, agents can handle more diverse and complex tasks, such as knowledge-based question answering~\citep{Search-o1, Search-R1, MMSearch-R1}, visual question answering(VQA)~\citep{DetToolChain, BDoG}, and visual grounding~\citep{HYDRA, UniVG-R1}.

In the training-based agent paradigm, supervised approaches such as supervised fine-tuning (SFT)~\citep{LLaVA-Plus} and reinforcement learning (RL)~\citep{Visual-ARFT, MMSearch-R1, Search-R1} have become the dominant strategies. However, these methods incur substantial resource costs, particularly in the multimodal domain, where high-quality data synthesis and model training are both expensive and labor-intensive.

In contrast, the training-free agent paradigm offers greater flexibility and lower deployment costs ~\citep{Cantor, IoT, VAP}. Yet, existing frameworks often fall into local reasoning traps, lacking long-horizon decision-making and coherent strategic planning~\citep{long-horizon}. These limitations arise from the absence of a global deliberative module that can provide high-level guidance across extended reasoning trajectories. Furthermore, most current methods remain predominantly text-centric, overlooking the distinctive demands of multimodal cognition—in particular, the need to preserve visual context and maintain situational awareness across iterative perception-action loops. Some approaches~\citep{Cantor, IoT} attempt to mitigate this by incorporating standalone visual foundation models as external tools, but this often increases systemic complexity and reduces usability and integration efficiency in practical deployments.

\begin{figure*}[t]
\centering
\includegraphics[width=1.0\textwidth]{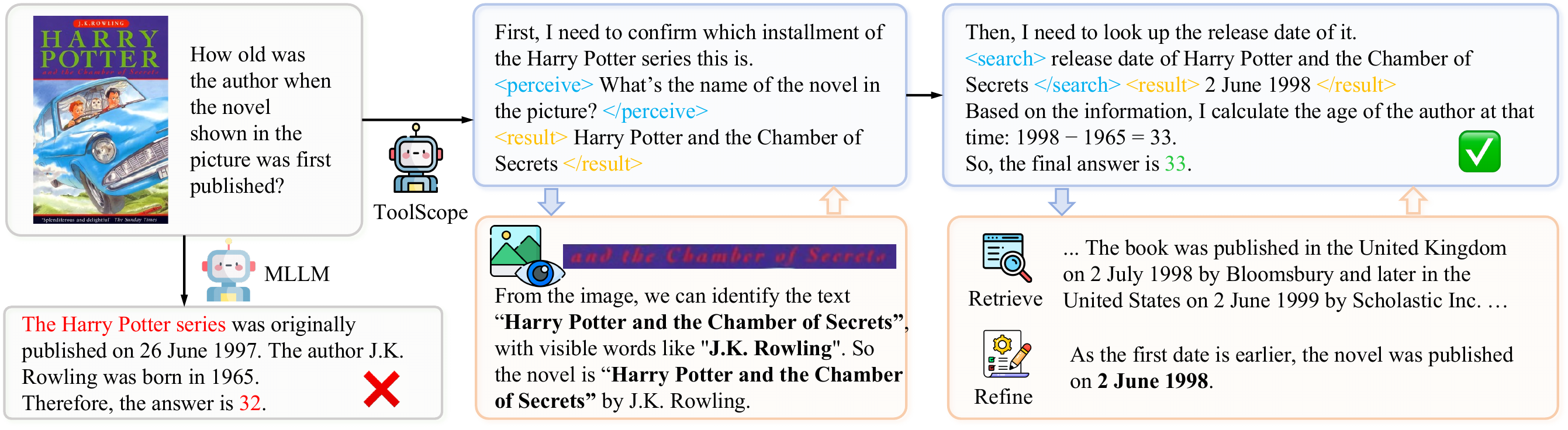}
\caption{The Illustration of the benefits of our \methodname\ to perform agentic tool-augmented reasoning in multimodal tasks. \methodname\ enable MLLMs to zoom in on the image in detail, retrieve external knowledge to augment reasoning.}
\label{comparison}
\end{figure*}

Despite their methodological differences, both training-based and training-free paradigms face two core challenges. \textbf{(1) Limited Global Planning.} These approaches predominantly rely on step-by-step decision-making, lacking mechanisms for global analysis and strategic planning. This often results in suboptimal tool selection and disjointed reasoning trajectories, particularly in tasks that require cross-modal interactions. \textbf{(2) Visual Context Degradation.} They exhibit limited capacity for visual context preservation over extended inference. Once visual information is initially processed, it is rarely retained or revisited in subsequent reasoning steps, leading to degraded performance. As illustrated in Figure~\ref{comparison}, the MLLM fails to attend to critical visual details, erroneously identifying the novel as the Harry Potter series rather than recognizing it as the second installment in it. 

To overcome these limitations, we aim to bridge this gap through following research questions:

\textbf{(1)} How can global planning be integrated with localized multimodal perception to enable more coherent and effective decision-making in multimodal agents?

\textbf{(2)} How can visual information be retained, referenced, and reused across multi-step inference to mitigate the problem of visual context degradation?

To address these limitations, we present \textbf{\methodname}, a multimodal agent framework that unifies global planning and local multimodal perception through three modular components: the Global Navigator, the Agentic Executor, and the Response Synthesizer. The Global Navigator performs high-level task decomposition and long-term tool selection, providing strategic guidance before any tool invocation. The Agentic Executor then conducts step-by-step reasoning, invoking tools in a visual grounded manner to solve decomposed subproblems. By tightly integrating top-down decision-making with bottom-up execution, \methodname\ achieves coherent, adaptive, and semantically aligned reasoning across complex multimodal tasks.

To address the challenge of visual context degradation over extended reasoning trajectories, \methodname\ incorporates a specialized perception module, the Perceive tool. This component enables dynamic visual grounding by allowing the agent to formulate and answer focused visual subquestions throughout the reasoning process. Unlike prior approaches that rely on static, one-time visual encoding, \methodname\ treats the image as a queryable perceptual memory, enabling selective re-attention to relevant regions as needed. This capability facilitates fine-grained perceptual grounding and improves visual consistency across long-horizon decision-making.

In our experiments on four diverse benchmarks of VQA 2.0, ScienceQA, MAT-Search, and MathVista, \methodname\ outperforms all baselines, achieving an average accuracy improvement of up to +6.69\%. Besides, \methodname\ achieves the highest average performance across 3 MLLM series of Qwen2.5-VL~\citep{Qwen2.5-VL}, InternVL3~\citep{InternVL3} and MiMo-VL~\citep{Mimo-VL}, demonstrating strong robustness and generalizability.

Our contributions are summarized as follows:
\begin{itemize}
    \item We propose \methodname, a modular multimodal agent framework comprising a Global Navigator, Agentic Executor, and Response Synthesizer, effectively combining global planning with local multimodal perception.
    
    \item \methodname\ autonomously invokes external tools, including a code executor, a multimodal retriever, and a perceiving module. The perception module facilitates the agent to re-attend to images and mitigates visual context degradation in long-horizon VQA tasks.
    
    \item We conduct extensive experiments on four VQA datasets spanning diverse domains. Results show that \methodname\ achieves consistent and significant improvements over baselines, demonstrating strong generalization ability.
    
    \item We provide a detailed analysis including ablation studies, qualitative analysis and case studies, underscoring the effectiveness of tool use in agentic multimodal reasoning.
\end{itemize}

\section{Related Work}


\textbf{Agentic Tool Use in MLLMs} Introducing agentic tool use into MLLMs offers a promising solution to key limitations including restricted internal knowledge and limited dynamic reasoning. 
Training-based approaches enable models to use tools through either supervised fine-tuning~\citep{LLaVA-Plus} or reinforcement learning~\citep{Visual-ARFT, MMSearch-R1}, but they often incur high resource costs—especially in multimodal domains.

Inspired by advances in language-only tool use, several studies~\citep{MM-REACT, Visual_ChatGPT} have extended the ReAct framework~\citep{ReAct} to the multimodal setting. These approaches integrate external visual experts, allowing models to tackle complex tasks by interleaving intermediate reasoning steps with tool calls.
Planning-and-execution approaches~\citep{IoT, Visual_Sketchpad} further advance this paradigm by enabling models to coordinate sequences of image operations. These operations are executed either by pretrained vision foundation models or external libraries such as Matplotlib and Pillow, with their results composited to produce a final answer.
Workflow-based approaches~\citep{Cantor} follow a more rigid design, guiding the MLLM to solve tasks through predefined, static workflows that invoke external visual tools or experts.
However, these existing methods often overlook global context or lack coherent integration between global and local perception. Moreover, the issue of visual context degradation remains underexplored. In this paper, we propose a training-free agentic framework that tightly integrates global and local perception and effectively mitigates visual context degradation through a dedicated perceive module.

\section{Preliminary}
\subsection{Problem Formulation}
We consider the problem of multimodal visual question answering (VQA) with tool-augmented reasoning. Given an input consisting of an image $I$ and a natural language question $Q$, the goal is to generate an accurate and grounded answer $A$, potentially through the use of external tools $T$. These tools can include vision experts, OCR engines, knowledge retrievers, code executors, and others.
Our proposed framework, \methodname, addresses this task by explicitly decomposing the reasoning process into three stages: global planning, local tool-augmented reasoning, and final response synthesis.
The overall generation probability of the reasoning chain $R$ and the answer $A$ is modeled as:
\begin{align}
P(A \mid I,Q,T)
&= \overbrace{P(G \mid I,Q,T)}^{\text{Global Planning}} \notag\\
&\quad \times 
    \overbrace{
    \prod_{s=1}^{S} P(R_s \mid I,Q,G,R_{<s})
    }^{\text{Agentic Execution}} \notag\\
&\quad \times 
    \underbrace{
    P(A \mid I,Q,R)}_{\text{Response Synthesizing}}.
\end{align}

In this formulation, $G$ denotes the global plan, comprising a selected subset of relevant tools and a global guidance that outlines the high-level strategy for task resolution.
The complete reasoning chain $R = { R_1, \dots, R_S }$ is generated iteratively through stepwise tool invocations or internal deductions, conditioned on the current context $(I, Q)$, the global plan $G$, and previous reasoning steps $R_{<s}$.
Finally, $A$ represents the final response synthesized from the reasoning trace.

\section{Methodology}

\begin{figure*}[t]
\centering
\includegraphics[width=1.0\textwidth]{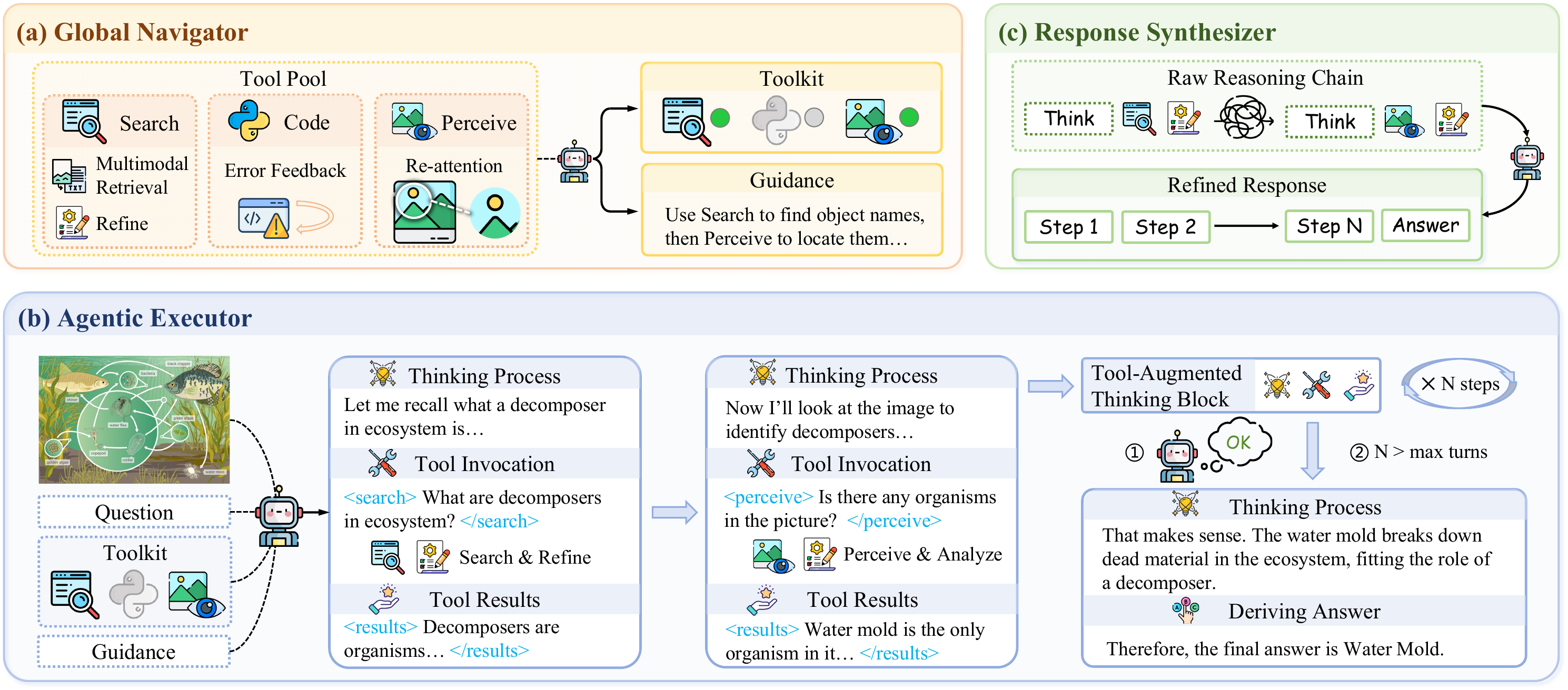}
\caption{\textbf{Overview of \methodname.} It consists of three components: \textbf{(a) Global Navigator} selects a subset toolkit from the tool pool, and generates global guidance. \textbf{(b) Agentic Executor} works iteratively to think, execute tool invocation, and continue reasoning based on tools. \textbf{(c) Response Synthesizer} consolidates the reasoning trajectory into a user-friendly response.}
\label{overview}
\end{figure*}

\subsection{Method Overview} 
Our proposed framework, \methodname, is a plug-and-play multimodal agent designed to solve complex VQA tasks through global planning, iterative multimodal tool use, and response synthesis. As shown in Figure~\ref{overview}, the overall framework consists of three components: Global Navigator, Agentic Executor, and Response Synthesizer, each responsible for a specific stage in the reasoning pipeline.

Formally, given a multimodal input $(I, Q)$ and predefined tools set $T$,  \methodname\ operates in the following three stages: 

\textbf{(1)} First of all, the Global Navigator analyzes the input $(I, Q, T)$ to provide a global planning $G$, consisting a selected subset toolkit and a high-level guidance, reducing unnecessary tool invocation complexity.

\textbf{(2)} Then, the Agentic Executor takes $(I, Q, G)$ as input and generates a tool-augmented reasoning trace $R$. It autonomously generates tool invocations, integrates returned results into the reasoning chain, and continues the reasoning process. This process may repeat multiple times to accommodate complex multi-step reasoning needs.

\textbf{(3)} Finally, the Response Synthesizer consolidates the reasoning trajectory $R$, filters out redundant or failed trials, and produces the final user-friendly response $A$.

In the following sections, we will describe each component in detail.

\subsection{Global Navigator}
\label{global_navigator}
The Global Navigator is responsible for high-level strategic planning and toolkit selection. Given a multimodal input $(I, Q)$, this module first performs a global analysis to understand the overall intent and complexity of the task, formulating a global reasoning strategy to guide the subsequent execution steps. 

The other key function of the Global Navigator is to select an appropriate subset of tools $T' \subseteq T$, where $T$ denotes the full tool pool. This selection plays a crucial role in reducing reasoning complexity, narrowing the model’s decision space, and enabling more focused tool use in the downstream reasoning. 
When the problem is simple and can be solved solely by the inherent capabilities of the MLLM, the Global Navigator may choose to not use any tools at all, i.e., setting $T' = \varnothing$. In this case, the agent functions without any external tool augmentation.
Conversely, for more compositional and complex tasks, the Global Navigator may select multiple tools to support different aspects of the reasoning pipeline. For instance, a question that requires both factual lookup and numerical computation may result in a toolkit $T' = \{\texttt{Search}, \texttt{Code}\}$, enabling the agent to decompose and solve subproblems in a modular way.

The output of the Global Navigator is a structured guidance prompt $G$ that encapsulates both a high-level reasoning plan $g$ and the selected tool subset $T'$, which are passed to the Agentic Executor to steer the subsequent tool-augmented reasoning process. This overall process is formulated as:
\begin{align}
    & P(G \mid I,Q,T) \notag \\
    & = P(T' \mid I, Q, T) \cdot \prod_{n=1}^{N} P(g_n \mid I, Q, T, g_{<n}).
\end{align}

\subsection{Agentic Executor}
\label{agentic_executor}
The Agentic Executor is the core reasoning engine that carries out multi-step problem solving in a tool-augmented, iterative manner. It performs autonomous tool invocation by generating special tool-call tokens during inference. Once a tool is invoked, its result is dynamically injected into the ongoing context, allowing the executor to continue reasoning based on the newly acquired information.

To tackle the challenges posed by diverse VQA tasks requiring different skills, MLLMs must be equipped with three fundamental capabilities:

\textbf{(1) Dynamic Visual Perception:} The competence to maintain visual grounding across multi-step reasoning, including re-examining specific image regions and adapting perception based on evolving context.

\textbf{(2) Multimodal Knowledge Acquisition:} The ability to access and integrate factual, commonsense, or image relevant information that is not contained within the model's internal parameters.

\textbf{(3) Numerical Computation Ability:} The capacity to perform precise arithmetic, logical inference, and symbolic computation, especially for questions involving measurements, equations, or structured data.

Therefore, we introduce three specialized tools that equip the agent with these core capabilities, each outlined in the following paragraphs.

\textbf{Perceive Tool} equips the agent with the ability to dynamically re-attend to the image during multi-step reasoning, addressing the common issue of visual context degradation. It enables the agent to formulate targeted visual sub-questions and extract localized visual information on demand. This mechanism not only enhances compositional reasoning over visual content but also unlocks MLLM's latent visual understanding by treating perception as an active and recursive process. Perceive is implemented natively by the agent’s backbone MLLM without any external detectors, OCR engines, or specialized perception models.

\textbf{Search Tool} supports both textual and multimodal retrieval to provide the agent with external knowledge beyond what is encoded in the base MLLM. (1) For textual retrieval, we employ a classic BM25 on Wikipedia dump. (2) For multimodal retrieval, we constructs a multimodal knowledge base using the training set of the target dataset as an image corpus. We employ CLIP-based cross-modal retrieval to identify relevant examples that may support the current inference step. We define the similarity function of the visual query $I$ and textual knowledge $t$ in Equation~\ref{eq:similarity}, where $f_{\text{image}}$ and $f_{\text{text}}$ represents the image encoding and text encoding module of CLIP~\citep{CLIP} respectively. To reduce noise and avoid introducing irrelevant visual information that could mislead reasoning, we apply a cosine similarity threshold of $\tau = 0.9$. Only images whose similarity to the query exceeds this threshold are retained, i.e $\text{sim}(I, t) > \tau$. This selective retrieval mechanism ensures that only semantically relevant visual content is introduced during reasoning, enhancing the robustness of the agent’s responses. 
\begin{equation}
    \text{sim}(I, t) = \cos ( f_{\text{image}} (I), f_{\text{text}} (t) )
    \label{eq:similarity}
\end{equation}

\begin{table*}[t] 
    \centering
    \footnotesize  
    \setlength{\tabcolsep}{11pt}{
        \begin{tabular}{llccccc}
		\toprule
            \textbf{Model} & \textbf{Method} & \textbf{VQA 2.0} & \textbf{ScienceQA} & \textbf{MAT-Search} & \textbf{MathVista} & \textbf{Avg.} \\
            \cmidrule{1-7}
            \multirow{7}*{Qwen2.5-VL-7B}
            & Direct Prompting & 66.33 & 82.00 & 33.33 & 61.7 & 60.84 \\
            & CoT & 55.00 & 82.67 & \underline{40.00} & \underline{61.9} & 59.89 \\
            & PoT & 50.67 & \underline{85.00} & 30.67 & 56.7 & 55.76 \\
            & Naïve MRAG & \underline{67.18} & 77.67 & 39.33 & 62.0 & \underline{61.54} \\
            & Cantor & 41.00 & 83.67 & 24.00 & 59.9 & 52.14 \\
            & \cellcolor{babyblue}\methodname\ & \cellcolor{babyblue}\textbf{71.18} & \cellcolor{babyblue}\textbf{87.97} & \cellcolor{babyblue}\textbf{40.67} & \cellcolor{babyblue}\textbf{62.8} & \cellcolor{babyblue}\textbf{65.65} \\

            \midrule
            \multirow{7}*{InternVL3-8B}
            & Direct Prompting & 70.33 & 88.00 & 33.33 & 68.0 & \underline{64.91} \\
            & CoT & 58.89 & \underline{88.67} & 38.82 & \underline{68.7} & 63.77 \\
            & PoT & 55.84 & 86.72 & 29.84 & 65.4 & 59.45 \\
            & Naïve MRAG & \underline{71.45} & 78.33 & \underline{40.26} & \underline{68.7} & 64.68 \\
            & Cantor & 58.12 & 88.33 & 27.33 & 68.4 & 60.54 \\
            & \cellcolor{babyblue}\methodname\ & \cellcolor{babyblue}\textbf{74.91} & \cellcolor{babyblue}\textbf{90.14} & \cellcolor{babyblue}\textbf{46.00} & \cellcolor{babyblue}\textbf{70.6} & \cellcolor{babyblue}\textbf{70.41} \\

            \midrule
            \multirow{6}*{MiMo-VL-7B-RL}
            & Direct Prompting & 68.67 & 83.33 & 34.67 & 73.3 & 64.99 \\
            & CoT & 60.17 & 84.02 & 44.39 & \underline{74.9} & 65.87 \\
            & PoT & 56.00 & 84.56 & 31.00 & 68.2 & 59.94 \\
            & Naïve MRAG & \underline{69.33} & 79.48 & \underline{46.97} & 73.8 & \underline{67.39} \\
            & Cantor & 46.82 & \underline{85.29} & 29.17 & 74.0 & 58.82 \\
            & \cellcolor{babyblue}\methodname\ & \cellcolor{babyblue}\textbf{70.07} & \cellcolor{babyblue}\textbf{88.00} & \cellcolor{babyblue}\textbf{52.45} & \cellcolor{babyblue}\textbf{76.2} & \cellcolor{babyblue}\textbf{71.68} \\

            \midrule
            \multirow{6}*{InternVL3-78B}
            & Direct Prompting & 74.81 & 91.67 & 35.68 & 75.6 & 69.44 \\
            & CoT & 68.20 & \underline{92.15} & 37.77 & 75.0 & 68.28 \\
            & PoT & 62.73 & 88.00 & 30.84 & 73.8 & 63.84 \\
            & Naïve MRAG &  \underline{76.34} & 80.63 & \underline{42.98} & \underline{75.9} & \underline{68.96} \\
            & Cantor & 71.92 & 90.48 & 28.00 & 74.0 & 66.10 \\
            & \cellcolor{babyblue}\methodname\ & \cellcolor{babyblue}\textbf{79.47} & \cellcolor{babyblue}\textbf{93.00} & \cellcolor{babyblue}\textbf{54.42} & \cellcolor{babyblue}\textbf{76.2} & \cellcolor{babyblue}\textbf{75.77} \\
            
		\bottomrule
		\end{tabular}
 	}
    \caption{\textbf{Main Results of \methodname on four benchmarks.}  The best scores are highlighted in \textbf{bold} and the second are marked in \underline{underline}.}
    \label{tab:main_result}
\end{table*}

\textbf{Code Tool} allows the agent to write and execute Python code for solving numerically intensive or algorithmically complex subproblems. Recognizing the limited code generation ability of even advanced MLLMs, we design a feedback loop where execution errors (e.g., syntax or runtime errors) are captured and fed back to the model. The model then revises the code based on the error trace until a valid result is obtained.

The Agentic Executor runs iteratively to perform step-by-step reasoning to obtain the full reasoning trace $R = \{R_1, R_2, ..., R_S\}$. At each step $s$, the reasoning trace $R_s$ consists of the thinking process $r$, tool invocation $q$, and the corresponding tool execution result $a$. Concretely, it first generates a thinking process $r$ deciding how to invoke external tools, then generates a tool invocation $q$, and calls the tool to solve the proposed query with an answer $a$. The overall process can be formulated as:
\begin{align}
    & P(R_s \mid I, Q, G, R_{<s})  \notag \\
    & = P_{\text{MLLM}}(r, q \mid I, Q, G, R_{<s}) 
    \times P_{\text{tool}}(a | I, Q, q)
\end{align}

\subsection{Response Synthesizer} 
\label{response_synthesizer}
After the iterative reasoning process, represented by the sequence of intermediate reasoning steps $R = \{ R_1, R_2, \dots, R_S \}$, the Response Synthesizer is responsible for aggregating these outputs and generating the final answer $A$. We formulate the probability of this process as $P(A \mid I, Q, R)$. The Response Synthesizer condenses the entire reasoning trajectory by filtering out redundant attempts, unsuccessful tool invocations, and irrelevant detours, thereby distilling the core steps that contribute to the correct solution. This module then refines the final response to ensure coherence, completeness, and faithful alignment with the original question $Q$ and image $I$. By performing this final synthesis, the agent significantly enhances the clarity and reliability of its output, which is crucial for practical applications requiring concise and interpretable answers.

\section{Experiment}

\subsection{Experimental Setups}

\textbf{Datasets and Metrics.} We evaluate \methodname\ on four diverse VQA benchmarks: 
\textbf{(1) VQA 2.0}~\citep{VQA2.0} is a large-scale, general visual question answering dataset; we randomly sample 300 questions from its validation set to evaluate general vision-language understanding.  
\textbf{(2) ScienceQA}~\citep{ScienceQA} focuses on scientific knowledge and visual reasoning; we sample 300 image-based questions from its test split to test domain-specific reasoning. 
\textbf{(3) MAT-Search}~\citep{Visual-ARFT} is a retrieval-based VQA dataset requiring external knowledge, containing 150 test examples; we conducted experiments on the hard split of 75 questions to demonstrate our method's capability in handling complex VQA tasks.
\textbf{(4) MathVista}~\citep{MathVista} is a comprehensive dataset for mathematical reasoning with visual inputs; we use its testmini subset containing 1,000 examples to assess deep mathematical reasoning. 
We use the official accuracy metric for MathVista and VQA 2.0, and Exact Match (EM) accuracy for the other three datasets.

\textbf{Baselines.} We compare \methodname\ with two categories of baselines. The first includes prompt-based methods that rely solely on in-context reasoning without external tools: 
\textbf{(1) Direct Prompting}, where the model answers questions directly; 
\textbf{(2) Chain-of-Thought (CoT)} prompting, which encourages step-by-step reasoning; 
\textbf{(3) Program-of-Thought (PoT)} prompting, which guides reasoning through pseudo-code. 
The second category consists of tool-augmented agents that leverage external capabilities: 
\textbf{(1) Naïve MRAG}, a basic multimodal retrieval-augmented generation pipeline;  
\textbf{(2) Cantor}~\citep{Cantor}, a multimodal agent that plans and calls vision experts.

\subsection{Main Experimental Results}

The main experimental results of \methodname, evaluated across four benchmark datasets, are summarized in Table~\ref{tab:main_result}. As shown in the table, we have the following key insights:

\textbf{\methodname\ consistently outperforms baselines.} Overall, \methodname\ consistently achieves superior performance relative to all baselines. For example, \methodname\ based on MiMo-VL-7B-RL obtains a performance gain of up to +9.12\% on MAT-Search, +4.67\% on ScienceQA, and +1.4\% on both VQA 2.0 and MathVista. These findings demonstrate the robustness and general applicability of \methodname\, which achieves high performance without relying on any task-specific fine-tuning.

\textbf{\methodname\ generalizes well across different backbone families and model sizes.}
We observe that \methodname\ maintains its performance advantage across different backbone families, including Qwen2.5-VL, InternVL3 and MiMo-VL series. \methodname\ achieves an average improvement of +6.69\% across all datasets on MiMo-VL-7B-RL, +6.33\% on InternVL3-78B, +5.50\% on InternVL3-8B, and +4.81\% on Qwen2.5-VL-7B. These findings indicate that its benefits are not tied to any specific architecture. Moreover, its effectiveness scales with model capacity. \methodname\ leverages the increasing ability of larger models without sacrificing stability, making it a scalable solution for tool-augmented multimodal reasoning.

\textbf{Existing baselines show inconsistent performance across datasets.}
Despite their theoretical strengths, existing baselines fail to achieve consistent improvements across the full spectrum of VQA tasks. Specifically, CoT and Cantor demonstrate moderate gains on reasoning-heavy datasets like MathVista and ScienceQA, but underperform significantly on datasets requiring visual grounding or external knowledge access, such as VQA 2.0 and MAT-Search. Naïve achieves performance improvement on most datasets but exhibits reduced performance on ScienceQA. These results indicate that existing approaches are domain-specific and struggle to balance generality with performance.

\subsection{Ablation Study}
We conduct an ablation study to evaluate the contribution of each key component in our \methodname\ framework with Qwen2.5-VL-7B on two benchmark datasets: MathVista and MAT-Search. The results, summarized in Table~\ref{tab:ablation}, lead to the following observations: 
\textbf{(1) The Search tool is the most impactful.} Removing it results in a substantial performance drop, particularly on MAT-Search, where accuracy falls from 39.40\% to 33.10\%. This underscores the importance of external information.
\textbf{(2) Each tool plays a valuable role.} Removing the Code tool causes the largest performance drop on MathVista (from 65.3\% to 64.0\%), confirming its relevance for quantitative reasoning. Excluding the Perceive tool also leads to a performance decline, though to a lesser extent.

\begin{table}[t]
\centering
    \begin{tabular}{lcc}
    \toprule
    Method & MathVista & MAT-Search \\
    
    \midrule
    \quad w/o Search          & 64.3 & 33.10 \\
    \quad w/o Code            & 64.0 & 37.50 \\
    \quad w/o Perceive        & 64.9 & 36.21 \\
    
    \midrule
    \textbf{\methodname} & \textbf{65.3} & \textbf{39.40} \\
    \bottomrule
    \end{tabular}
\caption{Ablation study with Qwen2.5-VL-7B on MathVista and MAT-Search.}
\label{tab:ablation}
\end{table}

\subsection{Scaling with Top-k Retrieval Documents.}

In this section, we conduct a retrieval scaling analysis to investigate how varying the number of retrieved documents (top-$k$) impacts performance on MAT-Search and ScienceQA benchmarks, using Qwen2.5-VL-7B as the backbone. 

\begin{figure}[t]
\centering
\includegraphics[width=1.0\columnwidth]{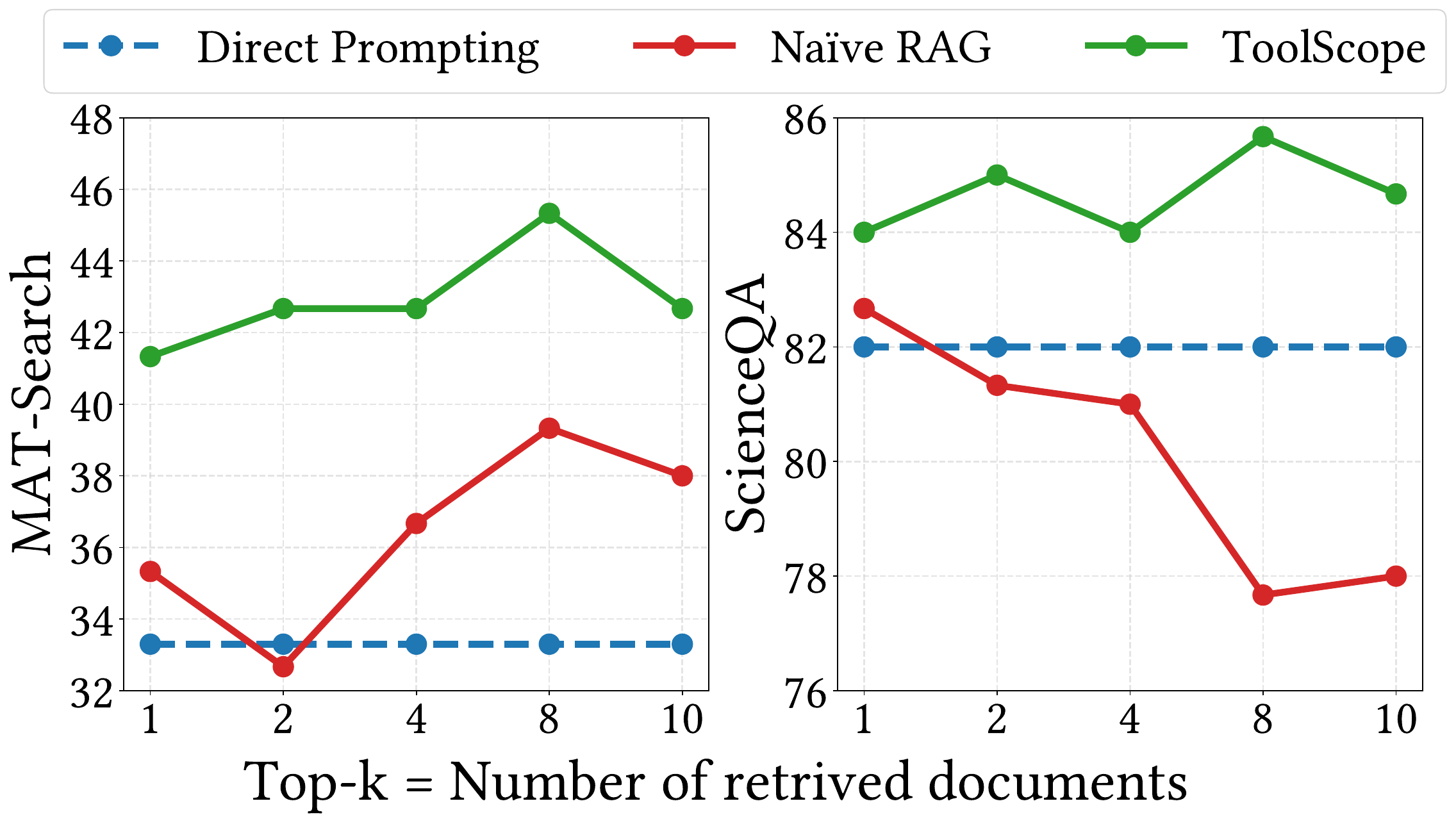}
\caption{Performance scaling with respect to the number of retrieved documents (top-k) using Qwen2.5-VL-7B. Results are shown on MAT-Search and ScienceQA. Increasing k allows access to more context but may introduce noise.}
\label{scaling_topk}
\end{figure}

From Figure~\ref{scaling_topk}, we draw the following conclusions:
\textbf{(1) \methodname\ consistently outperforms both baselines.} It achieves superior performance across all values of $k$, demonstrating its robustness in leveraging retrieved evidence effectively.
\textbf{(2) A clear trade-off emerges in the choice of top-$k$ values.} Selecting a small $k$ may fail to provide sufficient contextual information for effective reasoning, while a large $k$ introduces irrelevant or noisy content that can hinder performance. In our experiments, performance peaks at $k = 8$, suggesting this value strikes the optimal balance between informativeness and noise for the tasks considered.
\textbf{(3) Naïve RAG exhibits instability across datasets.} On MAT-Search, it shows a typical trade-off curve. However, on ScienceQA, performance deteriorates monotonically as $k$ increases, indicating early saturation due to irrelevant context and underscoring the method's limited robustness to noisy retrieval.

\subsection{Scaling with Reasoning Max Turns}

We conduct a scaling analysis to examine how varying the maximum number of reasoning turns influences model performance. Experiments are performed using Qwen2.5-VL-7B on MathVista and VQA 2.0, adopting Direct Prompting as baseline.

\begin{figure}[t]
\centering
\includegraphics[width=1.0\columnwidth]{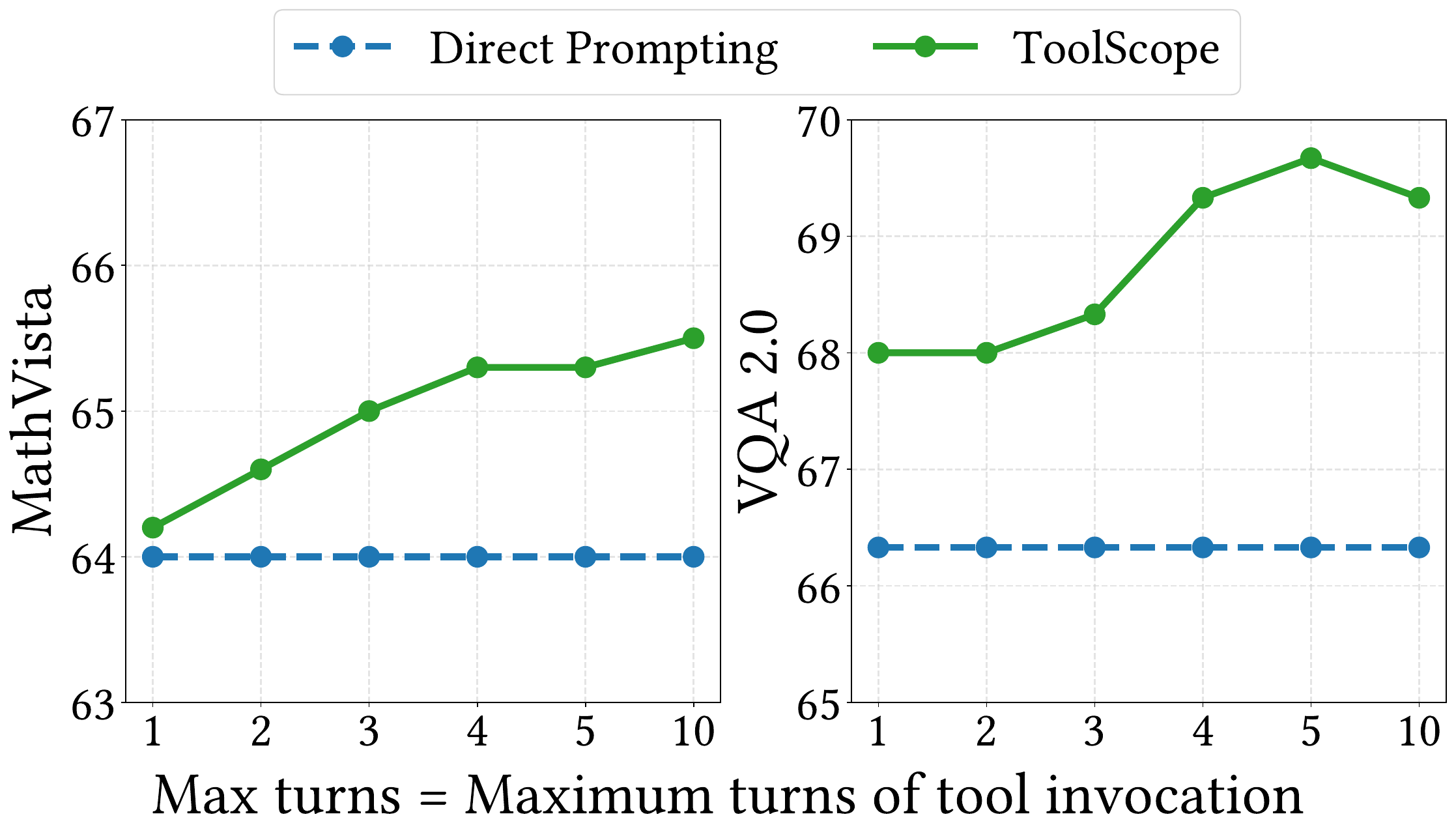} 
\caption{Scaling analysis of the maximum number of reasoning turns (max turns) using Qwen2.5-VL-7B on MathVista and VQA 2.0. }
\label{scaling_turns}
\end{figure}

From the results shown in Figure~\ref{scaling_turns}, we observe two key findings:
\textbf{(1) Increasing the maximum number of reasoning turns is effective and scalable.} Allowing more reasoning steps consistently improves the performance of \methodname\ on both datasets, highlighting the value of deeper, iterative reasoning and multi-step tool use in complex visual question answering tasks.
\textbf{(2) Performance gains exhibit diminishing returns.} The most significant improvements occur when increasing the maximum turns from 1 to 4. Beyond 5 turns, the incremental benefits taper off, indicating a marginal utility effect where additional steps contribute progressively less useful information.
\textbf{(3) Excessive reasoning depth may hinder performance on simpler tasks.} For datasets like VQA 2.0, which primarily involve shallow reasoning and minimal tool use, setting a large number of reasoning turns can lead to overthinking, potentially degrading performance. A moderate setting for max turns is thus preferable in such cases.

\subsection{Scaling with Model Size}

To investigate the impact of backbone model size on the performance, we conduct a scaling analysis on ScienceQA using five variants of the InternVL3 model: 2B, 8B, 9B, 14B and 78B parameters. 

\begin{figure}[t]
\centering
\includegraphics[width=1.0\columnwidth]{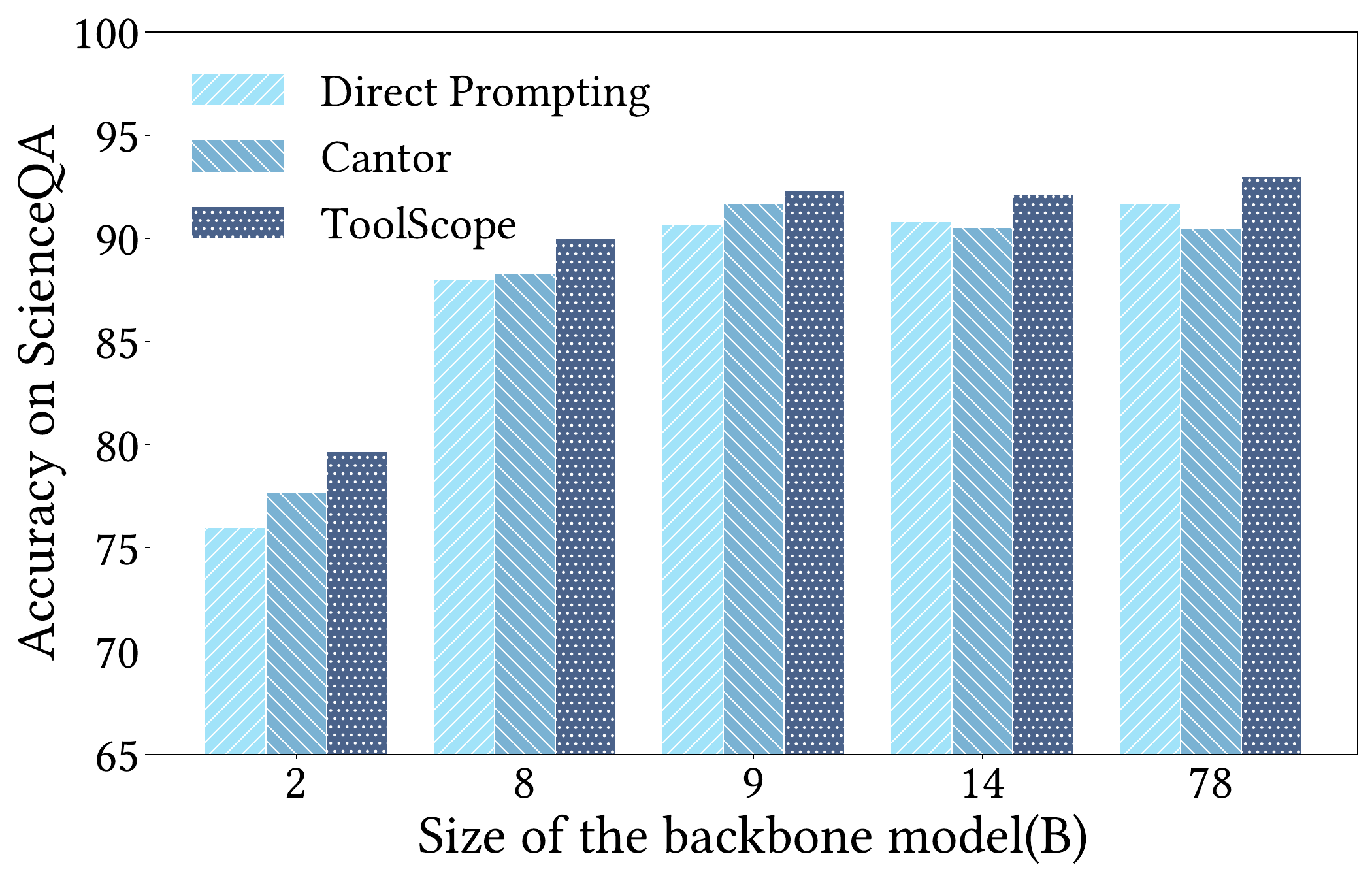}
\caption{Scaling analysis of the size of backbone models using InternVL3 series on ScienceQA.}
\label{scaling_size}
\end{figure}

As shown in Figure~\ref{scaling_size}, our key observations are: 
\textbf{(1) Our method consistently outperforms the baselines.} \methodname\ achieves superior performance across all model scales, demonstrating the robustness and adaptability of \methodname\ regardless of the underlying model capacity. Notably, even with larger backbones, our approach yields substantial gains, underscoring the effectiveness of tool-augmented reasoning and structured perception planning.
\textbf{(2) The method is highly scalable.} As model size increases, we observe a generally steady improvement in overall accuracy, indicating that models of larger capacity are better able to exploit tool use and achieve more effective global-local coordination.

\section{Conclusion}
In this work, we proposed \methodname\, a plug-and-play multimodal agent framework that integrates global task planning with local multimodal perception. 
\methodname\ integrates three key components—Global Navigator, Agentic Executor, and Response Synthesizer—to enable strategic planning, iterative multimodal tool-augmented reasoning, and final response refinement. 
To address the problem of visual context degradation, we further propose a dedicated Perceive tool, which treats the image as a dynamic perceptual memory, enabling the agent to formulate and resolve visual sub-questions on demand.
Without requiring any fine-tuning, \methodname\ demonstrates strong generalization across four diverse VQA benchmarks, outperforming both prompt-based and tool-based baselines. Our experiments show that the combination of global planning and multimodal tool usage significantly enhances the agent's reasoning capability and adaptability.
We believe \methodname\ provides a general and extensible blueprint for building more capable multimodal agents, and opens up promising directions for future research in multimodal tool-augmented reasoning.

\section{Limitations}

While \methodname\ is training-free and broadly applicable across backbones and tasks, several limitations remain. First, our evaluation focuses on four research-oriented benchmarks, which may not fully capture open-world or safety-critical scenarios. Real-user settings are left for future work. Second, although we compare against strong prompting/RAG baselines, the breadth of prior agentic frameworks is not exhaustively covered due to reproducibility and engineering constraints. Third, our retrieval experiments primarily use a Wikipedia knowledge source, extending to heterogeneous or domain-specific corpora may introduce new integration costs and failure modes. Finally, our Code tool targets short-running, stateless snippets and does not support long-running jobs, heavy dependencies, or file I/O without additional governance. Tool access (search and code execution) also raises ethical and safety considerations, such as hallucinated citations, unsafe code suggestions, or inadvertent leakage. Hence, stronger policy, auditing, and sandboxing are necessary for deployment beyond the research setting.

\bibliography{acl_latex}

\begin{thebibliography}{30}
\providecommand{\natexlab}[1]{#1}

\bibitem[{{Anthropic}(2024)}]{Claude3}
{Anthropic}. 2024.
\newblock \href {https://www.anthropic.com/news/claude-3-family} {The {Claude 3} model family: {Opus}, {Sonnet}, {Haiku}}.
\newblock Technical report, Anthropic.

\bibitem[{Bai et~al.(2025{\natexlab{a}})Bai, Chen, Liu, Wang, Ge, Song, Dang, Wang, Wang, Tang, Zhong, Zhu, Yang, Li, Wan, Wang, Ding, Fu, Xu, Ye, Zhang, Xie, Cheng, Zhang, Yang, Xu, and Lin}]{Qwen2.5-VL}
Shuai Bai, Keqin Chen, Xuejing Liu, Jialin Wang, Wenbin Ge, Sibo Song, Kai Dang, Peng Wang, Shijie Wang, Jun Tang, Humen Zhong, Yuanzhi Zhu, Mingkun Yang, Zhaohai Li, Jianqiang Wan, Pengfei Wang, Wei Ding, Zheren Fu, Yiheng Xu, and 8 others. 2025{\natexlab{a}}.
\newblock Qwen2.5-vl technical report.
\newblock \emph{arXiv preprint arXiv:2502.13923}.

\bibitem[{Bai et~al.(2025{\natexlab{b}})Bai, Li, Liu, Tang, Zhang, Sun, Chu, and Tang}]{UniVG-R1}
Sule Bai, Mingxing Li, Yong Liu, Jing Tang, Haoji Zhang, Lei Sun, Xiangxiang Chu, and Yansong Tang. 2025{\natexlab{b}}.
\newblock \href {https://doi.org/10.48550/ARXIV.2505.14231} {Univg-r1: Reasoning guided universal visual grounding with reinforcement learning}.
\newblock \emph{CoRR}, abs/2505.14231.

\bibitem[{Gao et~al.(2024)Gao, Chen, Zhang, Fu, Shen, Zhang, Zhang, Zheng, Sun, Cao, and Ji}]{Cantor}
Timin Gao, Peixian Chen, Mengdan Zhang, Chaoyou Fu, Yunhang Shen, Yan Zhang, Shengchuan Zhang, Xiawu Zheng, Xing Sun, Liujuan Cao, and Rongrong Ji. 2024.
\newblock \href {https://doi.org/10.1145/3664647.3681249} {Cantor: Inspiring multimodal chain-of-thought of {MLLM}}.
\newblock In \emph{Proceedings of the 32nd {ACM} International Conference on Multimedia, {MM} 2024, Melbourne, VIC, Australia, 28 October 2024 - 1 November 2024}, pages 9096--9105. {ACM}.

\bibitem[{Goyal et~al.(2017)Goyal, Khot, Summers{-}Stay, Batra, and Parikh}]{VQA2.0}
Yash Goyal, Tejas Khot, Douglas Summers{-}Stay, Dhruv Batra, and Devi Parikh. 2017.
\newblock \href {https://doi.org/10.1109/CVPR.2017.670} {Making the {V} in {VQA} matter: Elevating the role of image understanding in visual question answering}.
\newblock In \emph{2017 {IEEE} Conference on Computer Vision and Pattern Recognition, {CVPR} 2017, Honolulu, HI, USA, July 21-26, 2017}, pages 6325--6334. {IEEE} Computer Society.

\bibitem[{Hu et~al.(2024)Hu, Shi, Fu, Roth, Ostendorf, Zettlemoyer, Smith, and Krishna}]{Visual_Sketchpad}
Yushi Hu, Weijia Shi, Xingyu Fu, Dan Roth, Mari Ostendorf, Luke Zettlemoyer, Noah~A. Smith, and Ranjay Krishna. 2024.
\newblock \href {http://papers.nips.cc/paper\_files/paper/2024/hash/fb82011040977c7712409fbdb5456647-Abstract-Conference.html} {Visual sketchpad: Sketching as a visual chain of thought for multimodal language models}.
\newblock In \emph{Advances in Neural Information Processing Systems 38: Annual Conference on Neural Information Processing Systems 2024, NeurIPS 2024, Vancouver, BC, Canada, December 10 - 15, 2024}.

\bibitem[{Jin et~al.(2025)Jin, Zeng, Yue, Wang, Zamani, and Han}]{Search-R1}
Bowen Jin, Hansi Zeng, Zhenrui Yue, Dong Wang, Hamed Zamani, and Jiawei Han. 2025.
\newblock \href {https://doi.org/10.48550/ARXIV.2503.09516} {Search-r1: Training llms to reason and leverage search engines with reinforcement learning}.
\newblock \emph{CoRR}, abs/2503.09516.

\bibitem[{Ke et~al.(2024)Ke, Cai, Jahangard, Wang, Haghighi, and Rezatofighi}]{HYDRA}
Fucai Ke, Zhixi Cai, Simindokht Jahangard, Weiqing Wang, Pari~Delir Haghighi, and Hamid Rezatofighi. 2024.
\newblock \href {https://doi.org/10.1007/978-3-031-72661-3\_8} {{HYDRA:} {A} hyper agent for dynamic compositional visual reasoning}.
\newblock In \emph{Computer Vision - {ECCV} 2024 - 18th European Conference, Milan, Italy, September 29-October 4, 2024, Proceedings, Part {XX}}, volume 15078 of \emph{Lecture Notes in Computer Science}, pages 132--149. Springer.

\bibitem[{Li et~al.(2025)Li, Dong, Jin, Zhang, Zhou, Zhu, Zhang, and Dou}]{Search-o1}
Xiaoxi Li, Guanting Dong, Jiajie Jin, Yuyao Zhang, Yujia Zhou, Yutao Zhu, Peitian Zhang, and Zhicheng Dou. 2025.
\newblock \href {https://doi.org/10.48550/ARXIV.2501.05366} {Search-o1: Agentic search-enhanced large reasoning models}.
\newblock \emph{CoRR}, abs/2501.05366.

\bibitem[{Liu et~al.(2024)Liu, Cheng, Liu, Zhang, Li, Ren, Zou, Yang, Su, Zhu, Zhang, Gao, and Li}]{LLaVA-Plus}
Shilong Liu, Hao Cheng, Haotian Liu, Hao Zhang, Feng Li, Tianhe Ren, Xueyan Zou, Jianwei Yang, Hang Su, Jun Zhu, Lei Zhang, Jianfeng Gao, and Chunyuan Li. 2024.
\newblock \href {https://doi.org/10.1007/978-3-031-72970-6\_8} {Llava-plus: Learning to use tools for creating multimodal agents}.
\newblock In \emph{Computer Vision - {ECCV} 2024 - 18th European Conference, Milan, Italy, September 29-October 4, 2024, Proceedings, Part {XLVII}}, volume 15105 of \emph{Lecture Notes in Computer Science}, pages 126--142. Springer.

\bibitem[{Liu et~al.(2025)Liu, Zang, Zou, Liang, Dong, Cao, Duan, Lin, and Wang}]{Visual-ARFT}
Ziyu Liu, Yuhang Zang, Yushan Zou, Zijian Liang, Xiaoyi Dong, Yuhang Cao, Haodong Duan, Dahua Lin, and Jiaqi Wang. 2025.
\newblock \href {https://doi.org/10.48550/ARXIV.2505.14246} {Visual agentic reinforcement fine-tuning}.
\newblock \emph{CoRR}, abs/2505.14246.

\bibitem[{Lu et~al.(2024)Lu, Bansal, Xia, Liu, Li, Hajishirzi, Cheng, Chang, Galley, and Gao}]{MathVista}
Pan Lu, Hritik Bansal, Tony Xia, Jiacheng Liu, Chunyuan Li, Hannaneh Hajishirzi, Hao Cheng, Kai{-}Wei Chang, Michel Galley, and Jianfeng Gao. 2024.
\newblock \href {https://openreview.net/forum?id=KUNzEQMWU7} {Mathvista: Evaluating mathematical reasoning of foundation models in visual contexts}.
\newblock In \emph{The Twelfth International Conference on Learning Representations, {ICLR} 2024, Vienna, Austria, May 7-11, 2024}. OpenReview.net.

\bibitem[{Lu et~al.(2022)Lu, Mishra, Xia, Qiu, Chang, Zhu, Tafjord, Clark, and Kalyan}]{ScienceQA}
Pan Lu, Swaroop Mishra, Tanglin Xia, Liang Qiu, Kai{-}Wei Chang, Song{-}Chun Zhu, Oyvind Tafjord, Peter Clark, and Ashwin Kalyan. 2022.
\newblock \href {http://papers.nips.cc/paper\_files/paper/2022/hash/11332b6b6cf4485b84afadb1352d3a9a-Abstract-Conference.html} {Learn to explain: Multimodal reasoning via thought chains for science question answering}.
\newblock In \emph{Advances in Neural Information Processing Systems 35: Annual Conference on Neural Information Processing Systems 2022, NeurIPS 2022, New Orleans, LA, USA, November 28 - December 9, 2022}.

\bibitem[{Meng et~al.(2023)Meng, Yang, Wang, and Zhang}]{Chain-of-Image}
Fanxu Meng, Haotong Yang, Yiding Wang, and Muhan Zhang. 2023.
\newblock \href {https://doi.org/10.48550/ARXIV.2311.09241} {Chain of images for intuitively reasoning}.
\newblock \emph{CoRR}, abs/2311.09241.

\bibitem[{Moon et~al.(2024)Moon, Madotto, Lin, Nagarajan, Smith, Jain, Yeh, Murugesan, Heidari, Liu, Srinet, Damavandi, and Kumar}]{AnyMAL}
Seungwhan Moon, Andrea Madotto, Zhaojiang Lin, Tushar Nagarajan, Matt Smith, Shashank Jain, Chun{-}Fu Yeh, Prakash Murugesan, Peyman Heidari, Yue Liu, Kavya Srinet, Babak Damavandi, and Anuj Kumar. 2024.
\newblock \href {https://doi.org/10.18653/V1/2024.EMNLP-INDUSTRY.98} {Anymal: An efficient and scalable any-modality augmented language model}.
\newblock In \emph{Proceedings of the 2024 Conference on Empirical Methods in Natural Language Processing: {EMNLP} 2024 - Industry Track, Miami, Florida, USA, November 12-16, 2024}, pages 1314--1332. Association for Computational Linguistics.

\bibitem[{OpenAI(2023)}]{GPT-4V}
OpenAI. 2023.
\newblock {GPT-4V(ision) System Card}.
\newblock \url{https://cdn.openai.com/papers/GPT-4V(ision)_system_card.pdf}.
\newblock \url{https://cdn.openai.com/papers/GPT-4V(ision)_system_card.pdf}.

\bibitem[{Radford et~al.(2021)Radford, Kim, Hallacy, Ramesh, Goh, Agarwal, Sastry, Askell, Mishkin, Clark, Krueger, and Sutskever}]{CLIP}
Alec Radford, Jong~Wook Kim, Chris Hallacy, Aditya Ramesh, Gabriel Goh, Sandhini Agarwal, Girish Sastry, Amanda Askell, Pamela Mishkin, Jack Clark, Gretchen Krueger, and Ilya Sutskever. 2021.
\newblock \href {http://proceedings.mlr.press/v139/radford21a.html} {Learning transferable visual models from natural language supervision}.
\newblock In \emph{Proceedings of the 38th International Conference on Machine Learning, {ICML} 2021, 18-24 July 2021, Virtual Event}, volume 139 of \emph{Proceedings of Machine Learning Research}, pages 8748--8763. {PMLR}.

\bibitem[{Song et~al.(2025)Song, Chen, Liu, Chen, Li, and Lin}]{long-horizon}
Xinshuai Song, Weixing Chen, Yang Liu, Weikai Chen, Guanbin Li, and Liang Lin. 2025.
\newblock \href {https://openaccess.thecvf.com/content/CVPR2025/html/Song\_Towards\_Long-Horizon\_Vision-Language\_Navigation\_Platform\_Benchmark\_and\_Method\_CVPR\_2025\_paper.html} {Towards long-horizon vision-language navigation: Platform, benchmark and method}.
\newblock In \emph{{IEEE/CVF} Conference on Computer Vision and Pattern Recognition, {CVPR} 2025, Nashville, TN, USA, June 11-15, 2025}, pages 12078--12088. Computer Vision Foundation / {IEEE}.

\bibitem[{Team et~al.(2023)Team, Anil, Borgeaud, Alayrac, Yu, Soricut, Schalkwyk, Dai, Hauth, Millican et~al.}]{Gemini}
Gemini Team, Rohan Anil, Sebastian Borgeaud, Jean-Baptiste Alayrac, Jiahui Yu, Radu Soricut, Johan Schalkwyk, Andrew~M Dai, Anja Hauth, Katie Millican, and 1 others. 2023.
\newblock Gemini: a family of highly capable multimodal models.
\newblock \emph{arXiv preprint arXiv:2312.11805}.

\bibitem[{Wu et~al.(2023)Wu, Yin, Qi, Wang, Tang, and Duan}]{Visual_ChatGPT}
Chenfei Wu, Shengming Yin, Weizhen Qi, Xiaodong Wang, Zecheng Tang, and Nan Duan. 2023.
\newblock \href {https://doi.org/10.48550/ARXIV.2303.04671} {Visual chatgpt: Talking, drawing and editing with visual foundation models}.
\newblock \emph{CoRR}, abs/2303.04671.

\bibitem[{Wu et~al.(2025)Wu, Deng, Li, Liu, You, Li, Ma, and Liu}]{MMSearch-R1}
Jinming Wu, Zihao Deng, Wei Li, Yiding Liu, Bo~You, Bo~Li, Zejun Ma, and Ziwei Liu. 2025.
\newblock \href {https://doi.org/10.48550/ARXIV.2506.20670} {Mmsearch-r1: Incentivizing lmms to search}.
\newblock \emph{CoRR}, abs/2506.20670.

\bibitem[{Wu et~al.(2024)Wu, Wang, Tang, Wu, He, Ouyang, Torr, and Wu}]{DetToolChain}
Yixuan Wu, Yizhou Wang, Shixiang Tang, Wenhao Wu, Tong He, Wanli Ouyang, Philip Torr, and Jian Wu. 2024.
\newblock \href {https://doi.org/10.1007/978-3-031-73411-3\_10} {Dettoolchain: {A} new prompting paradigm to unleash detection ability of {MLLM}}.
\newblock In \emph{Computer Vision - {ECCV} 2024 - 18th European Conference, Milan, Italy, September 29-October 4, 2024, Proceedings, Part {XXXII}}, volume 15090 of \emph{Lecture Notes in Computer Science}, pages 164--182. Springer.

\bibitem[{Xiao et~al.(2024)Xiao, Zhang, Han, Fu, Yu, Zhong, Wu, Wang, Yin, and Chen}]{VAP}
Ziyang Xiao, Dongxiang Zhang, Xiongwei Han, Xiaojin Fu, Wing~Yin Yu, Tao Zhong, Sai Wu, Yuan Wang, Jianwei Yin, and Gang Chen. 2024.
\newblock \href {http://papers.nips.cc/paper\_files/paper/2024/hash/328c922d068dd4ccb23cec5c64e6c7fc-Abstract-Conference.html} {Enhancing {LLM} reasoning via vision-augmented prompting}.
\newblock In \emph{Advances in Neural Information Processing Systems 38: Annual Conference on Neural Information Processing Systems 2024, NeurIPS 2024, Vancouver, BC, Canada, December 10 - 15, 2024}.

\bibitem[{Xiaomi(2025)}]{Mimo-VL}
LLM-Core-Team Xiaomi. 2025.
\newblock \href {https://arxiv.org/abs/2506.03569} {Mimo-vl technical report}.
\newblock \emph{Preprint}, arXiv:2506.03569.

\bibitem[{Yang et~al.(2025)Yang, Tan, Wu, Zheng, Peng, Liang, Gu, Cai, Ye, Jang, Deng, and Gao}]{Magma}
Jianwei Yang, Reuben Tan, Qianhui Wu, Ruijie Zheng, Baolin Peng, Yongyuan Liang, Yu~Gu, Mu~Cai, Seonghyeon Ye, Joel Jang, Yuquan Deng, and Jianfeng Gao. 2025.
\newblock \href {https://openaccess.thecvf.com/content/CVPR2025/html/Yang\_Magma\_A\_Foundation\_Model\_for\_Multimodal\_AI\_Agents\_CVPR\_2025\_paper.html} {Magma: {A} foundation model for multimodal {AI} agents}.
\newblock In \emph{{IEEE/CVF} Conference on Computer Vision and Pattern Recognition, {CVPR} 2025, Nashville, TN, USA, June 11-15, 2025}, pages 14203--14214. Computer Vision Foundation / {IEEE}.

\bibitem[{Yang et~al.(2023)Yang, Li, Wang, Lin, Azarnasab, Ahmed, Liu, Liu, Zeng, and Wang}]{MM-REACT}
Zhengyuan Yang, Linjie Li, Jianfeng Wang, Kevin Lin, Ehsan Azarnasab, Faisal Ahmed, Zicheng Liu, Ce~Liu, Michael Zeng, and Lijuan Wang. 2023.
\newblock \href {https://doi.org/10.48550/ARXIV.2303.11381} {{MM-REACT:} prompting chatgpt for multimodal reasoning and action}.
\newblock \emph{CoRR}, abs/2303.11381.

\bibitem[{Yao et~al.(2023)Yao, Zhao, Yu, Du, Shafran, Narasimhan, and Cao}]{ReAct}
Shunyu Yao, Jeffrey Zhao, Dian Yu, Nan Du, Izhak Shafran, Karthik~R. Narasimhan, and Yuan Cao. 2023.
\newblock \href {https://openreview.net/forum?id=WE\_vluYUL-X} {React: Synergizing reasoning and acting in language models}.
\newblock In \emph{The Eleventh International Conference on Learning Representations, {ICLR} 2023, Kigali, Rwanda, May 1-5, 2023}. OpenReview.net.

\bibitem[{Zheng et~al.(2024)Zheng, Liang, Zhang, Wei, Chua, and Li}]{BDoG}
Changmeng Zheng, Dayong Liang, Wengyu Zhang, Xiaoyong Wei, Tat{-}Seng Chua, and Qing Li. 2024.
\newblock \href {https://doi.org/10.1145/3664647.3681102} {A picture is worth a graph: {A} blueprint debate paradigm for multimodal reasoning}.
\newblock In \emph{Proceedings of the 32nd {ACM} International Conference on Multimedia, {MM} 2024, Melbourne, VIC, Australia, 28 October 2024 - 1 November 2024}, pages 419--428. {ACM}.

\bibitem[{Zhou et~al.(2024)Zhou, Zhou, Hu, Lu, Gao, and Zhang}]{IoT}
Qiji Zhou, Ruochen Zhou, Zike Hu, Panzhong Lu, Siyang Gao, and Yue Zhang. 2024.
\newblock \href {https://doi.org/10.48550/ARXIV.2405.13872} {Image-of-thought prompting for visual reasoning refinement in multimodal large language models}.
\newblock \emph{CoRR}, abs/2405.13872.

\bibitem[{Zhu et~al.(2025)Zhu, Wang, Chen, Liu, Ye, Gu, Tian, Duan, Su, Shao, Gao, Cui, Wang, Cao, Liu, Wei, Zhang, Wang, Xu, Li, Wang, Deng, Li, He, Jiang, Luo, Wang, He, Shi, Zhang, Shao, He, Xiong, Qu, Sun, Jiao, Lv, Wu, Zhang, Deng, Ge, Chen, Wang, Dou, Lu, Zhu, Lu, Lin, Qiao, Dai, and Wang}]{InternVL3}
Jinguo Zhu, Weiyun Wang, Zhe Chen, Zhaoyang Liu, Shenglong Ye, Lixin Gu, Hao Tian, Yuchen Duan, Weijie Su, Jie Shao, Zhangwei Gao, Erfei Cui, Xuehui Wang, Yue Cao, Yangzhou Liu, Xingguang Wei, Hongjie Zhang, Haomin Wang, Weiye Xu, and 32 others. 2025.
\newblock \href {https://doi.org/10.48550/ARXIV.2504.10479} {Internvl3: Exploring advanced training and test-time recipes for open-source multimodal models}.
\newblock \emph{CoRR}, abs/2504.10479.

\end{thebibliography}

\clearpage
\appendix

\section{Appendix}
\label{sec:appendix}

\subsection{Implementation Details}
All experiments are performed with off-the-shelf multimodal LLMs in a strictly training-free setting. The Perceive capability is realized by the backbone models themselves; no external detectors, OCR engines, or specialized perception modules are attached. We cap the agent’s reasoning budget at 10 turns. In our setting, increasing this cap yields monotonically improved accuracy with only moderate latency growth, and a 10-turn budget offers a favorable accuracy–latency trade-off. 

Decoding hyperparameters are held fixed across all experiments. Qwen2.5-VL and MiMo-VL use temperature 0.7, top-p 0.9, top-k 50, and a repetition penalty of 1.05. InternVL3 uses temperature 0.8, top-p 0.8, top-k 40, and the same repetition penalty. Generation terminates on the model’s end-of-sequence token together with task-specific control tokens for tool invocation (</search>, </perceive>, </code>). Stop strings are retained in the returned outputs for auditing.

For retrieval experiments, we employ an English Wikipedia dump as the corpus. We remove extremely short pages (fewer than 32 words) and treat the remainder as clean documents. The clean set is segmented into passages with a chunk size of 256 and an overlap of 32, using a whitespace-based length function. This process yields approximately 15.8 billion passages in total. We use CLIP with the ViT-B/16 backbone as the retriever.

\subsection{Case Study}

\begin{figure}[t]
\centering
\includegraphics[width=1.0\columnwidth]{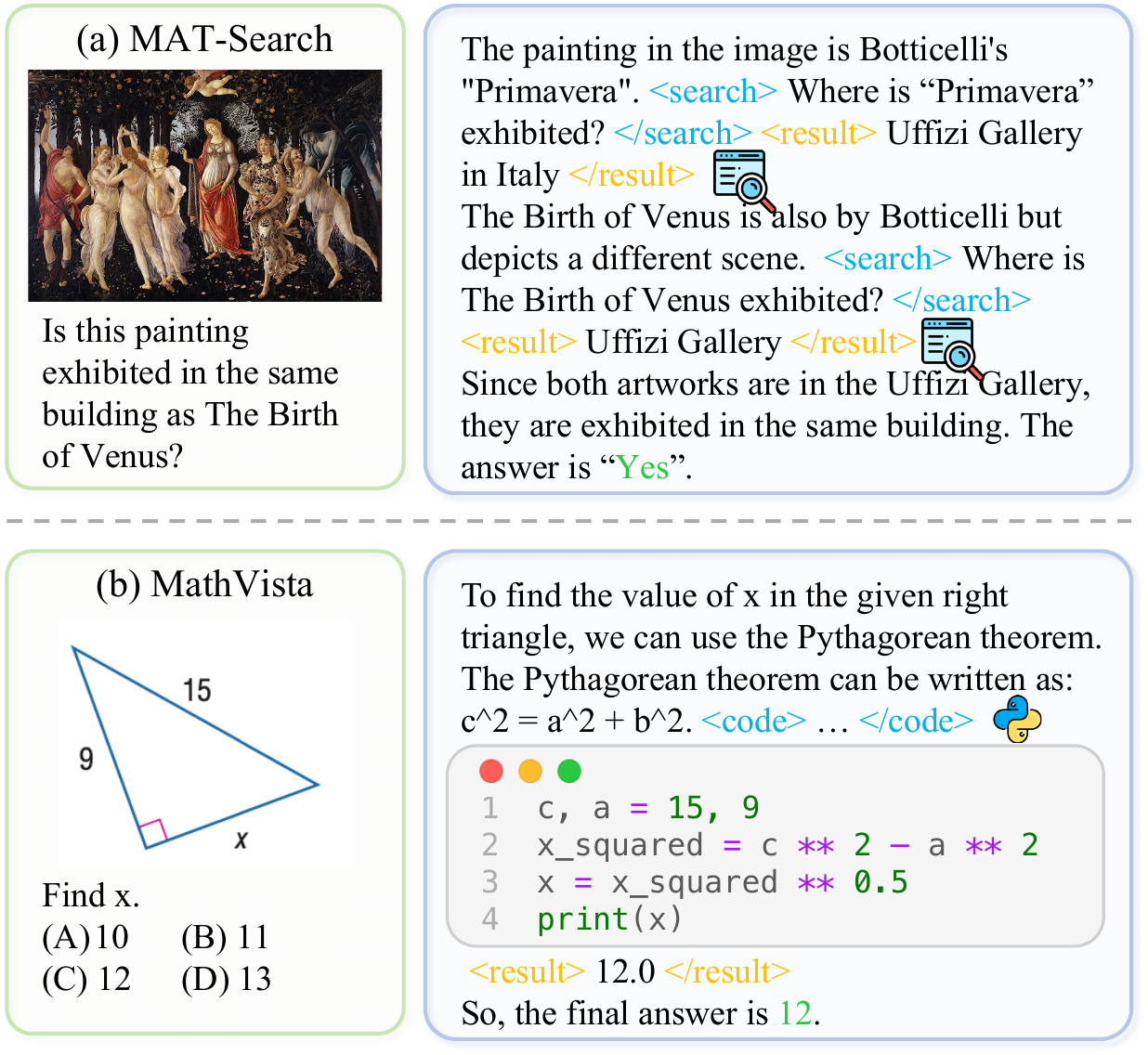}
\caption{A case study of MAT-Search and MathVista.}
\label{case_study}
\end{figure}

We present a case study featuring two representative examples, sourced from the MAT-Search and MathVista datasets, respectively. As illustrated in Figure~\ref{case_study}, for the MAT-Search example, \methodname\ autonomously selects the Search tool to retrieve the exhibition locations of two paintings. In contrast, for the MathVista example, \methodname\ chooses the Code tool to compute an unknown value using the Pythagorean theorem. 
These cases demonstrate \methodname's ability to intelligently select task-appropriate tools and employ them in a goal-directed, agentic manner tailored to the specific demands of each problem.

\subsection{Latency Analysis}

We measure end-to-end latency as the wall-clock time from receiving a question to producing the final answer. To reduce the influence of outliers, we report the median (p50) over the full evaluation set for each benchmark. We further separate MLLM time from overall latency to quantify the model’s contribution. All experiments are conducted on a single NVIDIA A800 with Qwen2.5-VL-7B (batch size $=1$; max turns $=10$).

Table~\ref{tab:latency_analysis} summarizes latency and LLM time alongside the observed average number of turns. We make the following observations:
(1) Despite a generous 10-turn cap, actual reasoning depth remains modest (3.29--4.21 turns), suggesting effective self-termination and low redundancy in tool use.
(2) Latency is shaped by both the LLM and tools, but LLM time is the principal bottleneck (approximately 65--71\% of p50), indicating that improvements to decoding efficiency are likely to yield the largest gains.
(3) Harder, multi-step tasks (MathVista, MAT-Search) naturally incur longer latencies due to deeper reasoning and, for MAT-Search, tighter coupling with retrieval.
(4) Overall, the median latencies of 2.3--5.9s are acceptable for research evaluation and interactive analysis, while leaving clear headroom for systems optimizations.

\begin{table}[t]
\centering
\setlength{\tabcolsep}{4.5pt}
\begin{tabular}{l c c c}
\toprule
\textbf{Dataset} & \textbf{Turns} & \textbf{Latency} & \textbf{MLLM Time} \\
\midrule
VQA2.0     & 3.61 & 2.3 & 1.5 \\
ScienceQA  & 3.29 & 3.4 & 2.3 \\
MathVista  & 4.23 & 5.9 & 4.2 \\
MAT-Search & 4.21 & 5.1 & 3.6 \\
\bottomrule
\end{tabular}
\caption{Latency Analysis.}
\label{tab:latency_analysis}
\end{table}

\subsection{Prompt}

\clearpage

\begin{figure*}[t]
\centering
\begingroup\hypersetup{linkcolor=white}
\begin{prompt}{Global Navigator\hfill{}(\S\ref{global_navigator})}{pr:global_navigator}
You are an expert planner in a multimodal reasoning system. Your role is to perform high-level strategic analysis and select the most appropriate tools to help solve a given question about an image.\\
\\
Your objectives are:\\
1. Tool Selection: From the available tools listed below, identify which are necessary to solve the task.\\
2. Global Reasoning Plan: Write a concise, high-level plan that outlines the sequence of major reasoning steps needed to arrive at the correct answer. This plan should reflect how the selected tools will be used and in what order.\\
\\
Available Tools:\\
1. Search: Retrieve factual or background knowledge relevant to the question.\\
2. Code: Perform mathematical computations or logical operations.\\
3. Perceive: Extract fine-grained visual information (e.g., text, objects, layout) from the image.\\
\\
Guidelines:\\
1. If the task is simple and solvable directly by the model, you may decide to use no tools.\\
2. For complex tasks, combine tools modularly to handle different reasoning needs.\\
3. Be explicit about why each tool is selected and how it contributes to the reasoning plan.\\
\\
Output Format:\\
Return your answer in JSON format with two keys: "selected\_tools" and "global\_plan".\\
Example:\\
\{\{\\
"selected\_tools": ["Search", "Code"],\\
"global\_plan": "First, use Search to retrieve background knowledge about the scientific concept in the question. Then, use Code to compute the final result based on the retrieved and extracted information."\\
\}\}\\
\\
Question: \\
\{\textbf{question}\}\\

\end{prompt}
\endgroup
\end{figure*}

\begin{figure*}[t]
\centering
\begingroup\hypersetup{linkcolor=white}
\begin{prompt}{Agentic Executor\hfill{}(\S\ref{agentic_executor})}{pr:agentic_executor}
You are an advanced question-answering agent equipped with specialized modules to aid in analyzing and responding to queries about images:\\
1. Search: This module performs searches on Wikipedia to gather relevant information for any query. It's especially useful for retrieving knowledge, definitions and problem-solving strategies. When you need this module, state your request as: "<search> your query or topic of interest </search>". If you want to use the image as query, state your request as "<search> image </search>".\\
2. Perceive: This module can percerive visual content to answer simple questions about an image. It is especially useful for answering visual sub-questions such as identifying objects, counting instances, estimating attributes (e.g., age, color, size), and describing spatial relationships. Use this module when reasoning requires detailed understanding of the image. When you need this module, state your request as: "<perceive> your question </perceive>".\\
3. Code: This module allows you to write and execute Python code to perform tasks such as calculations, data analysis, simulations, or solving algorithmic problems. It's ideal when a task requires logic implementation and mathematical computation. When you need this module, state your Python code as: "<code> your Python code </code>". Ensure your code is complete, syntactically correct, and uses proper Python naming conventions.\\

When faced with a question about an image, your task is to: \\
1. Reason step by step.\\
2. Utilize modules during your reasoning. When you need Search module, use <search> to request a search and end with </search>. When you need Perceive module, use "<perceive> your question </perceive>". When you need Code module, use <code> to run Python code and end with </code>.\\
3. Give the final answer. \\

Here are some examples:\\
\{\textbf{tool\_examples}\}

Please refer to the prompts and examples above to help me solve the following problem. \\
Question: \\
\{\textbf{question}\} \\

Answer: \\
\{\textbf{previous\_reasoning}\}

\end{prompt}
\endgroup
\end{figure*}

\begin{figure*}[t]
\centering
\begingroup\hypersetup{linkcolor=white}
\begin{prompt}{Response Synthesizer\hfill{}(\S\ref{response_synthesizer})}{pr:response_synthesizer}
You will be given a question, an image, and a solution generated from previous steps. Your task is to form a clear reasoning path, extract the final answer from the reasoning result and reformat it for evaluation. \\

Give the final answer directly. If you are given Options, your answer should be one of them. Otherwise, your answer should be very brief. \\

Question: \\
\{\textbf{question}\}\\

Answer: \\
\{\textbf{reasoning}\}

\end{prompt}
\endgroup
\end{figure*}

\begin{figure*}[t]
\centering
\begingroup\hypersetup{linkcolor=white}
\begin{prompt}{Refiner of Search Tool\hfill{}(\S\ref{agentic_executor})}{pr:refiner_of_search_tool}
You are an advanced reasoning agent. Given the Question about an image, the Previous Reasoning Steps, a Current Search Query, and a set of Searched Documents, your task is to:\\
1. Analyze the Previous Reasoning Steps and Current Search Query to understand the assistant's current objective and what specific information is required.\\
2. Read Searched Documents and identify content directly relevant to the Current Search Query.\\
3. Integrate and rephrase the extracted information smoothly into the reasoning chain. Do not quote or copy verbatim. Instead, answer the Current Search Query using fluent, natural language. Phrase the information as internal knowledge or informed commentary, using expressions like "According to external sources", "As is known", or "As the search results show".\\
\\
Here is an example:\\
Question:\\
Find $x$ so that each quadrilateral is a parallelogram.\\
\\
Previous Reasoning Steps:\\
The image shows a quadrilateral that resembles a parallelogram. The left side is 2x-5, and the right side is 3x-18. To determine the value of x that makes the quadrilateral a parallelogram, we need to use the property of a parallelogram to create an equation.\\
\\
Current Search Query:\\
Properties of parallelograms\\
\\
Searched Documents:\\
Passage 1: \\
A simple (non-self-intersecting) quadrilateral is a parallelogram if and only if any one of the following statements is true:\\
Two pairs of opposite sides are parallel (by definition).\\
Two pairs of opposite sides are equal in length.\\
Two pairs of opposite angles are equal in measure.\\
The diagonals bisect each other.\\
One pair of opposite sides is parallel and equal in length.\\
Adjacent angles are supplementary.\\
\\
Output:\\
In a parallelogram, opposite sides are equal.\\
\\
Now complete the task for the input below:\\
Question:\\
\{\textbf{question}\}\\
\\
Previous Reasoning Steps:\\
\{\textbf{previous\_reasoning}\}\\
\\
Current Search Query:\\
\{\textbf{calling}\}\\
\\
Searched Documents:
\{\textbf{raw\_result}\}\\
\\
Output:\\

\end{prompt}
\endgroup
\end{figure*}

\end{document}